\documentclass[lettersize,journal]{IEEEtran}
\usepackage{amsmath,amsfonts}
\usepackage{algorithmic}
\usepackage{algorithm}
\usepackage{array}
\usepackage[caption=false,font=normalsize,labelfont=sf,textfont=sf]{subfig}
\usepackage{textcomp}
\usepackage{stfloats}
\usepackage{url}
\usepackage{verbatim}
\usepackage{graphicx}
\graphicspath{{figs/}}
\usepackage{cite}
\usepackage{siunitx}
\usepackage{booktabs}
\usepackage{multirow}
\usepackage{mathtools}
\usepackage{rotating}
\begin{document}

\title{Generative Model-Based Attack on Learnable Image Encryption for Privacy-Preserving Deep Learning}

\author{AprilPyone~MaungMaung and Hitoshi~Kiya
}

\maketitle

\begin{abstract}
In this paper, we propose a novel generative model-based attack on learnable image encryption methods proposed for privacy-preserving deep learning. Various learnable encryption methods have been studied to protect the sensitive visual information of plain images, and some of them have been investigated to be robust enough against all existing attacks. However, previous attacks on image encryption focus only on traditional cryptanalytic attacks or reverse translation models, so these attacks cannot recover any visual information if a block-scrambling encryption step, which effectively destroys global information, is applied. Accordingly, in this paper, generative models are explored to evaluate whether such models can restore sensitive visual information from encrypted images for the first time. We first point out that encrypted images have some similarity with plain images in the embedding space. By taking advantage of leaked information from encrypted images, we propose a guided generative model as an attack on learnable image encryption to recover personally identifiable visual information. We implement the proposed attack in two ways by utilizing two state-of-the-art generative models: a StyleGAN-based model and latent diffusion-based one. Experiments were carried out on the CelebA-HQ and ImageNet datasets. Results show that images reconstructed by the proposed method have perceptual similarities to plain images.
\end{abstract}

\begin{IEEEkeywords}
Image Encryption, Image Security, Cryptanalysis.
\end{IEEEkeywords}

\section{Introduction}
\IEEEPARstart{I}{n} the information age, distributed systems are indispensable for processing information. Cloud computing has been integrated in many applications from work to personal life. Social networking services have become a trend. Even more, the metaverse is coming to popularity. Therefore, a lot of multimedia data are generated every second. With amazing technologies that consume multimedia data on the rise, privacy and security are in demand more than ever.

Secure and efficient communication of multimedia data requires both compression and encryption~\cite{zhou2013designing,chuman2019encryption}. For encryption, full encryption with provable security (such as RSA and AES) is the most secure option. However, many multimedia applications have been seeking a trade-off between security and other features such as low processing demands, tolerance of data loss, bitstream compliance, compatibility with machine learning, and signal processing in the encrypted domain. Therefore, perceptual encryption methods have been studied and developed to balance the trade-off~\cite{wang2020color,tang2015efficient,ito2009one,kurihara2015encryption,chuman2019encryption}.

In the context of secure image transmission, the traditional way is to use a Compression-then-Encryption (CtE) system. However, on social networking services (SNS) and cloud photo sharing services (CPSS), it is required to (multiply) re-compress uploaded images. To be in line with such services, it is preferred to use an Encryption-then-Compression (EtC) system~\cite{chuman2019encryption}. Moreover, researchers have proposed new perceptual image encryption methods (so-called learnable image encryption) that can be used in deep learning~\cite{2018-ICCETW-Tanaka,2019-Access-Warit,madono2020block}. A recent study also shows that existing block scrambling-based EtC images can also be classified with an isotropic network~\cite{maung2022privacy,zheng2022privacy}. With deep learning on the rise, learnable image encryption methods are attractive because such encrypted images can be used in deep learning models while preserving privacy without significant performance degradation. Therefore, evaluating the security of learnable encrypted images is paramount, especially for privacy-critical scenarios such as medical image analysis.

Previous attacks show that two learnable encryption methods~\cite{2018-ICCETW-Tanaka,2019-Access-Warit} are vulnerable to chosen-plaintext and ciphertext-only attacks~\cite{chang2020attacks}. In addition, deep neural network-based attacks, using a GAN-based approach~\cite{madono2021gan} or inverse transformation network-based approach~\cite{ito2021image}, are effective at restoring visual information from encrypted images. Nevertheless, these existing attack methods do not work if a block-scrambling operation is used in the encryption procedure~\cite{madono2021gan}.

Therefore, in this paper, we propose a generative model-based attack. The idea is that instead of reconstructing the exact plain images from encrypted ones, the attack aims to reconstruct the styles. We implement the proposed attack with two methods by utilizing the StyleGAN encoder~\cite{richardson2021encoding} and a latent diffusion model~\cite{rombach2022high} to realize the proposed attack. The first method is a GAN-based method in which we use the StyleGAN encoder~\cite{richardson2021encoding} and a pre-trained StyleGAN2 model~\cite{karras2020analyzing}. The second method is a state-of-the-art latent diffusion model~\cite{rombach2022high} with a pre-trained CLIP (Contrastive Language-Image Pre-Training~\cite{radford2021learning}) model. A part of this work (the first method) was introduced in~\cite{maung2022attack}. In this paper, we extend the idea of the first method as a general generative model-based attack and conduct additional experiments. We make the following contributions in this paper.
\begin{itemize}
 \item The proposed attack is the first in which generative models are used to attack encrypted images.
 \item We demonstrate two ways of implementing the proposed method: a StyleGAN-based method and a diffusion-based one.
 \item We conduct extensive experiments and show that learnable encrypted images are vulnerable if the encryption algorithm is known.
\end{itemize}
In experiments, the proposed attack is confirmed to outperform state-of-the-art attacks, especially for reconstructing encryption-then-compression (EtC) images.

The rest of this paper is structured as follows. Section~\ref{sec:related-work} presents related work on privacy-preserving deep learning, learnable image encryption, previous attacks, the StyleGAN encoder, and latent diffusion models. Section~\ref{sec:threat} defines the threat model used in this paper. Section~\ref{sec:proposed} puts forward the proposed attack. Section~\ref{sec:experiment} presents experiments and discussion, and Section~\ref{sec:conclusion} concludes this paper.

\section{Related Work\label{sec:related-work}}
Generally, privacy-preserving machine learning addresses three issues: the (1) privacy of datasets, (2) privacy of models, and (3) privacy of models' outputs~\cite{shokri2015privacy}. To address privacy issues, researchers have proposed various solutions.

\subsection{Cryptographic Methods}
Fully Homomorphic Encryption (FHE) allows users to do computation on encrypted data while preserving the ability to decrypt the corresponding computation~\cite{brakerski2014leveled,gentry2009fully,chillotti2020tfhe}. In contrast, Secure Multi-Party Computation (SMPC) requires two or more parties to jointly compute a function while keeping their inputs secret between parties~\cite{chaum1988multiparty,cho2018secure}. Another cryptographic approach is Trusted Execution Environments (TEE), which is explored only for classical machine learning algorithms~\cite{law2020secure}. Although cryptographic methods offer strong security guarantees, they are still computationally expensive~\cite{shokri2015privacy} or cannot be directly applied to deep learning models.

\subsection{Federated Learning and Differential Privacy}
Federated learning allows us to train a global model in a distributed manner in which only gradient updates are uploaded to a central server~\cite{kmyrsb16,fereidooni2021safelearn,xu2019verifynet}. However, in many settings, data owners may not cooperate, and there is no aggregating server. In addition, federated learning can protect privacy only during the training phase. When a model is deployed in an untrusted cloud server, there is no privacy for the test data.

Differential Privacy (DP) provides information-theoretic privacy guarantees and can be implemented in deep neural network training~\cite{abadi2016deep,subramani2021enabling,bu2021fast}. However, there is a severe trade-off between accuracy and privacy in DP-based solutions.

\subsection{Learnable Image Encryption}
\begin{figure}[!t]
\centering
\includegraphics[width=\linewidth]{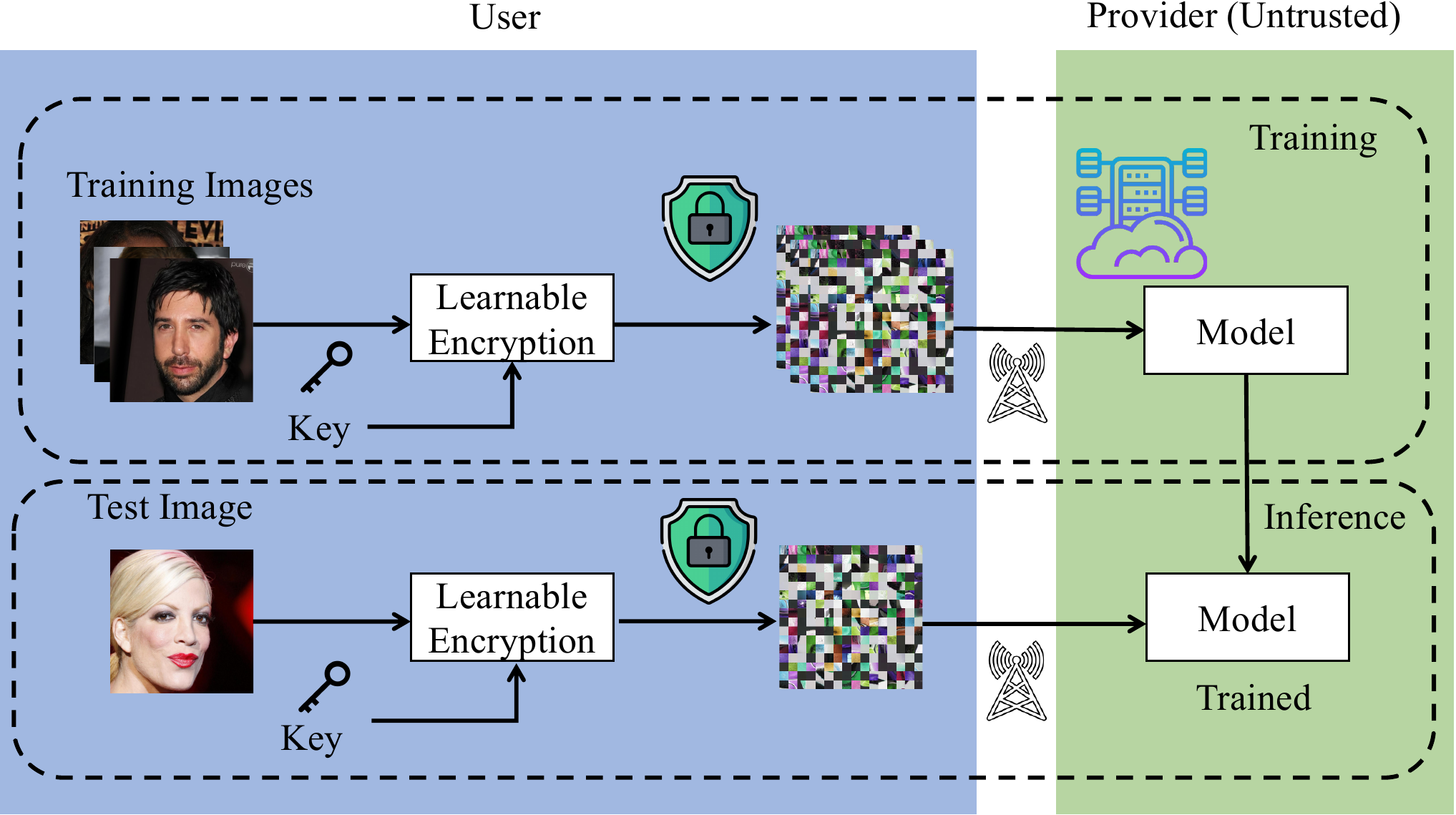}
\caption{Scenario of privacy-preserving image classification.\label{fig:classification-flow}}
\end{figure}

To overcome the limitations of the above privacy/security guaranteed methods, various learnable perceptual encryption methods, which have been inspired by encryption methods for privacy-preserving photo cloud sharing services~\cite{chuman2019encryption}, have been studied so far for various applications~\cite{kiya2022overview} such as image classification~\cite{2018-ICCETW-Tanaka,madono2020block,2019-Access-Warit}, semantic segmentation~\cite{kiya2022privacy}, adversarial defense~\cite{2020-Arxiv-Maung}, and model protection~\cite{maung2021protection,ito2022access,nagamori2023access}. Learnable image encryption is primarily designed for privacy-preserving image classification. A scenario involving privacy-preserving image classification is shown in Fig.~\ref{fig:classification-flow}, where a user encrypts training and test images before sending them to an untrusted cloud server (provider) for training and inference. The provider does not have the secret encryption key nor any visual information on encrypted images.

Learnable image encryption is a class of perceptual image encryption that is specifically designed for compatibility with deep neural networks. In other words, learnable encryption encrypts images with a secret key so that visual information in encrypted images is not perceptible to humans while maintaining the ability to classify encrypted images with a deep learning model. Tanaka first introduced block-wise learnable image encryption (LE) that utilizes a block-wise pixel shuffling operation~\cite{2018-ICCETW-Tanaka}. There is also a pixel-wise encryption (PE) approach in which negative-positive transformation and color component shuffling are applied~\cite{2019-Access-Warit}. However, both block-wise~\cite{2018-ICCETW-Tanaka} and pixel-wise~\cite{2019-Access-Warit} encryption methods can be attacked by ciphertext-only attacks~\cite{chang2020attacks}. To enhance the security of encryption, Tanaka's method (denoted as extended LE:\@ ELE) was extended by adding a block scrambling step and utilizing different block keys for the pixel encryption operation~\cite{madono2020block}.

\begin{figure}[!t]
\centering
\includegraphics[width=2.5in]{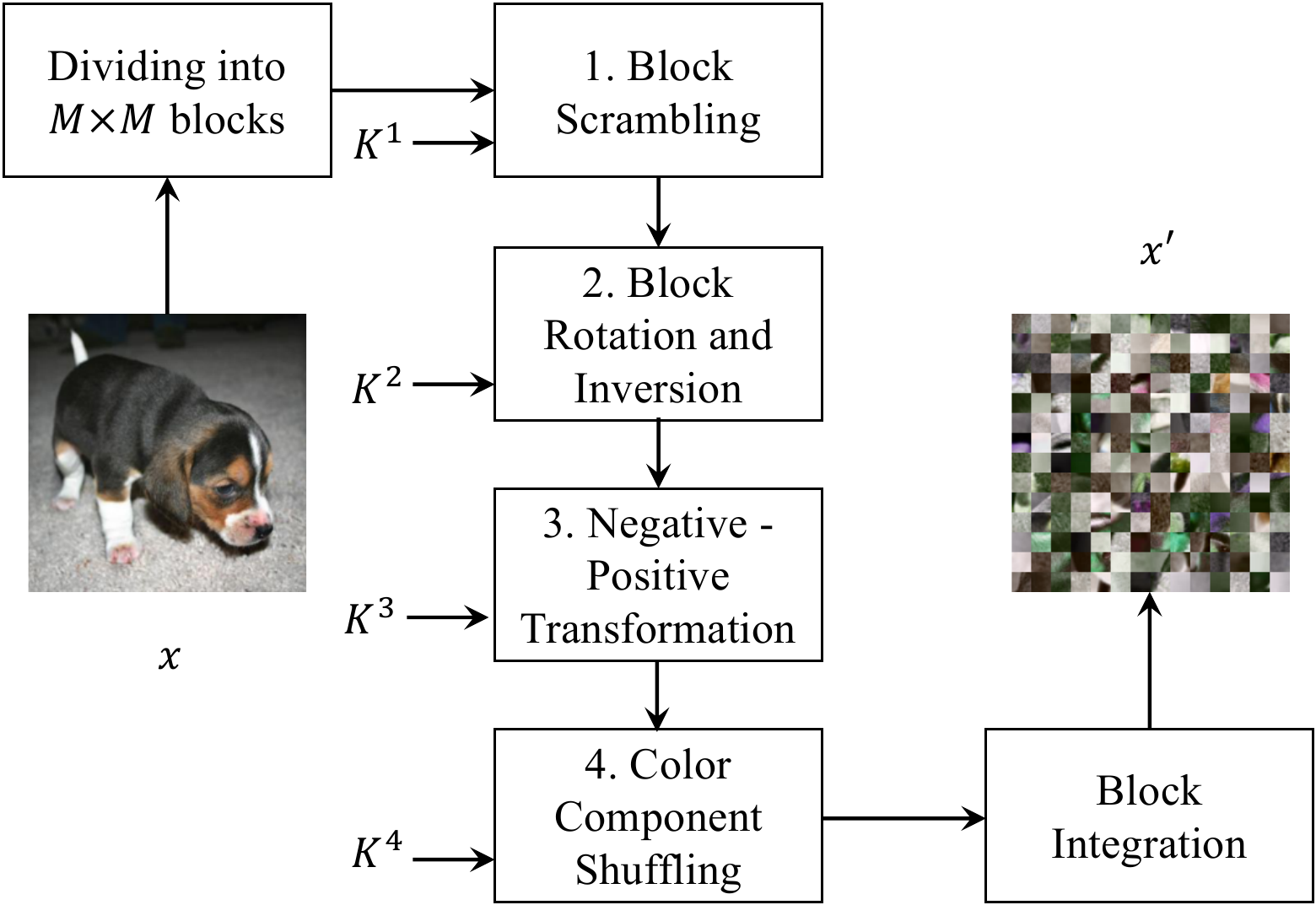}
\caption{Generation of EtC images.\label{fig:etc-procedure}}
\end{figure}

Moreover, with state-of-the-art deep neural networks, color-based Encryption-then-Compression (EtC) images~\cite{kurihara2015encryption,watanabe2015encryption,kurihara2015encryption2,sirichotedumrong2018grayscale} can also be classified~\cite{maung2022privacy}. We review the procedure of EtC encryption as described in Fig.~\ref{fig:etc-procedure}. A three-channel (RGB) color image ($x$) with $w \times h$ pixels is divided into non-overlapping blocks each with $M \times M$. Then, four encryption steps are carried out on the divided blocks as follows.
\begin{enumerate}
 \item Randomly permute the divided blocks by using a random integer generated by a secret key $K^1$.
 \item Rotate and invert each block randomly by using a random integer generated by a key $K^2$.
 \item Apply negative-positive transformation to each block by using a random binary integer generated by a key $K^3$, where $K^3$ is commonly used for all color components. A transformed pixel value in the $i$\textsuperscript{th} block, $p'$, is calculated using
 \begin{equation}
 p' = \left\{
 \begin{array}{ll}
 p & (r(i) = 0)\\
 p \oplus (2^L - 1) & (r(i) = 1),
 \end{array}
 \right.
 \end{equation}
where $r(i)$ is a random binary integer generated by $K^3$, $p$ is the pixel value of the original image with $L$ bits per pixel ($L=8$ is used in this paper), and $\oplus$ is the bitwise exclusive-or operation. The value of the occurrence probability $\mathrm{P}(r(i)) = 0.5$ is used to invert bits randomly.
\item Shuffle three color components in each block by using an integer randomly selected from six integers generated by a key $K^4$.
\end{enumerate}
Then, integrate the encrypted blocks to form an encrypted image $x'$. Note that block size $M = 8$ or $16$ is enforced to be JPEG compatible.

\begin{figure}[t]
\centering
\subfloat[]{\includegraphics[width=0.2\linewidth]{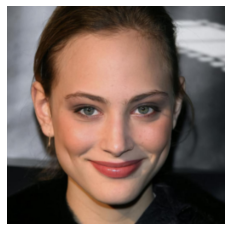}%
\label{fig:plain}}
\hfil
\subfloat[]{\includegraphics[width=0.2\linewidth]{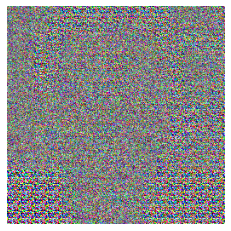}%
\label{fig:le}}
\hfil
\subfloat[]{\includegraphics[width=0.2\linewidth]{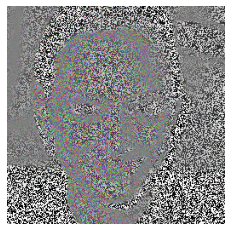}%
\label{fig:pe}}
\hfil
\subfloat[]{\includegraphics[width=0.2\linewidth]{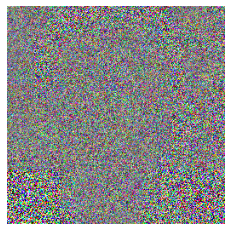}%
\label{fig:ele}}
\hfil
\subfloat[]{\includegraphics[width=0.2\linewidth]{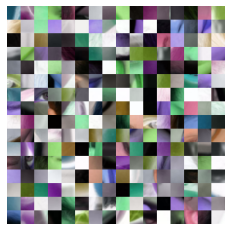}%
\label{fig:etc}}
\caption{Example of encrypted images. (a) Plain. (b) LE~\cite{2018-ICCETW-Tanaka}. (c) PE~\cite{2019-Access-Warit}. (c) ELE~\cite{madono2020block}. (d) EtC~\cite{kurihara2015encryption}.\label{fig:ex-encryption}}
\end{figure}

An example of encrypted images for different learnable image encryption methods is shown in Fig.~\ref{fig:ex-encryption}.

\subsection{Previous Attacks\label{sec:previous-attacks}}
Attack methods aim to restore visual information from encrypted images. There are two kinds of attack methods: the traditional cryptanalytic approach and the neural network-based one.

\subsubsection{Cryptanalytic Attack}
Previously, traditional chosen-plaintext and ciphertext-only attacks were proposed against LE and PE~\cite{chang2020attacks}. For EtC, encrypted images are special because they have almost the same correlation among pixels in each block as that of the original image for efficient compression. By exploiting this correlation property, an automatic jigsaw puzzle solver can be used as an attack~\cite{chuman2017icassp,chuman2017icme,chuman2018security}. These solver attack methods utilize pairwise compatibility and pairwise comparison, and it has been confirmed that assembling encrypted images is difficult if the number of blocks is large, the block size is small, and there is less color information~\cite{chuman2019encryption}. In addition, encrypted images that contain compression distortion (e.g., JPEG distortion) are much more difficult to decrypt with automatic jigsaw puzzle solvers.

\subsubsection{Neural Network-Based Attacks}
Some attacks use neural networks. The inverse transformation network attack utilizes a deep convolutional neural network to decrypt encrypted images by using plain and encrypted image pairs~\cite{ito2021image}. Another recent neural network-based attack is the generative adversarial network-based (GAN) attack~\cite{madono2021gan}. Similar to non-neural network-based attacks, neural network-based attacks have also been successfully applied to LE and PE.\@ However, since a block scrambling operation is used as in EtC and ELE, no attacks can reconstruct any visual information from encrypted images as in~\cite{madono2021gan}.

\subsection{StyleGAN Encoder}
The StyleGAN encoder was originally proposed for a general image-to-image translation task~\cite{richardson2021encoding}. This encoder encodes images into the latent space of a pre-trained StyleGAN generator. The encoder is trained by using a weighted combination of pixel-wise $\mathcal{L}_2$ loss, perceptual loss, regularization loss for latent vectors, and facial image recognition loss~\cite{richardson2021encoding}. We refer interested readers to the original paper~\cite{richardson2021encoding}. We use the StyleGAN encoder to extract styles from encrypted images in this paper.

\subsection{Latent Diffusion Models}
Denoising diffusion probabilistic models (DDPM) have shown state-of-the-art results in image generation~\cite{dhariwal2021diffusion}. DDPM is a latent variable model that consists of a forward process (gradually adding Gaussian noise to a sample from a true data distribution, $x_0 \sim q(x_0)$) and a reverse process (gradually denoising Gaussian noise to a true data sample with learned Gaussian transitions, $p_\theta(x_{t-1}|x_t)$)~\cite{sohl2015deep,ho2020denoising}. Both processes are defined as Markov chains. By following the formulations and notations in~\cite{ho2020denoising}, the forward process for a variance schedule $\beta_1, \ldots, \beta_T$ is
\begin{equation}
 q(x_t|x_{t-1}) \coloneqq \mathcal{N}(x_t; \sqrt{1 - \beta_t}x_{t-1}, \beta_t\boldsymbol{\mathrm{I}}).
\end{equation}
A convenient property is that a noised sample $x_t$ at an arbitrary timestep $t$ can be expressed in a closed form as
\begin{equation}
 \begin{split}
 & q(x_t|x_0) \coloneqq \mathcal{N}(x_t; \sqrt{\bar{\alpha}_t}x_0, (1 - \bar{\alpha}_t)\boldsymbol{\mathrm{I}}),\\
 & x_t = \sqrt{\bar{\alpha_t}}x_0 + \sqrt{(1 - \bar{\alpha}_t)}\epsilon,
\end{split}
\end{equation}
where $\alpha_t \coloneqq 1 - \beta_t$, $\bar{\alpha}_t \coloneqq \prod_{i=1}^{t}\alpha_i$, and $\epsilon \sim \mathcal{N}(0, \boldsymbol{\mathrm{I}})$. The DDPM in~\cite{ho2020denoising} utilizes a U-Net to predict $\epsilon_\theta(x_t, t)$ by using a standard mean squared error (MSE) loss. The simplified objective is
\begin{equation}
 \mathcal{L_{\text{(DDPM)}}} \coloneqq \mathbb{E}_{t \sim [1, T], x_0 \sim q(x_0), \epsilon \sim \mathcal{N}(0, \boldsymbol{\mathrm{I}})}[\lVert\epsilon - \epsilon_{\theta}(x_t, t)\rVert^2].
\end{equation}

Despite the great performance of diffusion models in generative modeling, diffusion models in pixel space consume an extensive amount of computational power. Therefore, latent diffusion models (LDM) have been proposed that worked in the latent space of a pre-trained autoencoder~\cite{rombach2022high}. The new objective for LDM is
\begin{equation}
 \mathcal{L_{\text{(LDM)}}} \coloneqq \mathbb{E}_{t \sim [1, T], \mathcal{E}(x_0) \sim q(\mathcal{E}(x_0)), \epsilon \sim \mathcal{N}(0, \boldsymbol{\mathrm{I}})}[\lVert\epsilon - \epsilon_{\theta}(z_t, t)\rVert^2],
\end{equation}
where $z_t$ is obtained from the encoder $\mathcal{E}$. The success of LDM can be seen in the current trending text-to-image model, known as \textbf{stable diffusion}. In this paper, we adopt LDM~\cite{rombach2022high} that is trained to reconstruct visual information from encrypted images.

\section{Threat Models\label{sec:threat}}
As we focus on the privacy of datasets in an image classification scenario, the goal of an adversary is to recover visual information on encrypted images. Encrypted images are transferred to an untrusted cloud provider for storage and training/testing a model as in Fig.~\ref{fig:classification-flow}. Therefore, conventional attack methods for learnable image encryption assume that the adversary has access to encrypted images and knows the encryption algorithm but not the secret key~\cite{chang2020attacks}. In other words, the adversary is assumed to carry out a ciphertext-only attack (COA) only from encrypted images. In this paper, we also assume that an attacker knows the plain-image data distribution and prepares a plain-encrypted paired dataset with an assumed key. Specifically, we define a threat model that includes a set of assumptions such as an attacker's goal, knowledge, and capabilities as follows.

\subsection{Adversary's Goals}
Given an encrypted image (cipher-image), the goal of an adversary is to recover some personally identifiable visual information. For example, if an encrypted image is a face image, the adversary intends to know the gender, hair color, skin color, eyeglasses, etc.

\subsection{Adversary's Knowledge}
We assume the adversary has full knowledge of the encryption algorithm but not the key. In addition, we also assume the adversary knows the plain-image data distribution. Therefore, the adversary can prepare a plain-encrypted paired dataset with a random key. Unlike the traditional COA, the plain-image data distribution is also available for the adversary in this paper.

\subsection{Adversary's Capabilities}
The adversary has access to GPU-accelerated computing resources. With the plain-encrypted paired dataset, the adversary can train a generative model.

\section{Proposed Attack\label{sec:proposed}}
Generally, an image-to-image translation model learns a mapping, $T : A \rightarrow B$, to translate images from domain $A$ to domain $B$. Mapping $T$ can be learned by generative adversarial networks either in a supervised~\cite{isola2017image} or unsupervised manner~\cite{zhu2017unpaired}. We aim to learn a mapping to translate an image from an encrypted domain to a plain domain. Nevertheless, learning such a mapping is challenging for two reasons:
\begin{enumerate}
 \item the encrypted domain and the plain domain are not aligned,
 \item there may be many encrypted images with different keys that correspond to one single plain image.
\end{enumerate}

Another approach to the image-to-image translation task is to utilize generative models with conditioning. In this paper, we propose an attack method that utilizes a generative model to recover some visual information from encrypted images for the first time. We demonstrate that by using encrypted images as conditional information, state-of-the-art generative models can reconstruct personal information such as gender, facial features, accessories, etc.

\subsection{Attack Formulation}
We hypothesize that encrypted images that are learnable by conventional deep neural networks have some relationship with plain images. The relationship between encrypted images and plain ones may cause some information to leak. We exploit this leaked information to recover visual features from encrypted images. We observe two properties in all conventional learnable image encryption methods. Let $x$ be a plain RGB image and $x'$ be an encrypted RGB image.

\begin{figure}[!t]
\centering
\includegraphics[width=\linewidth]{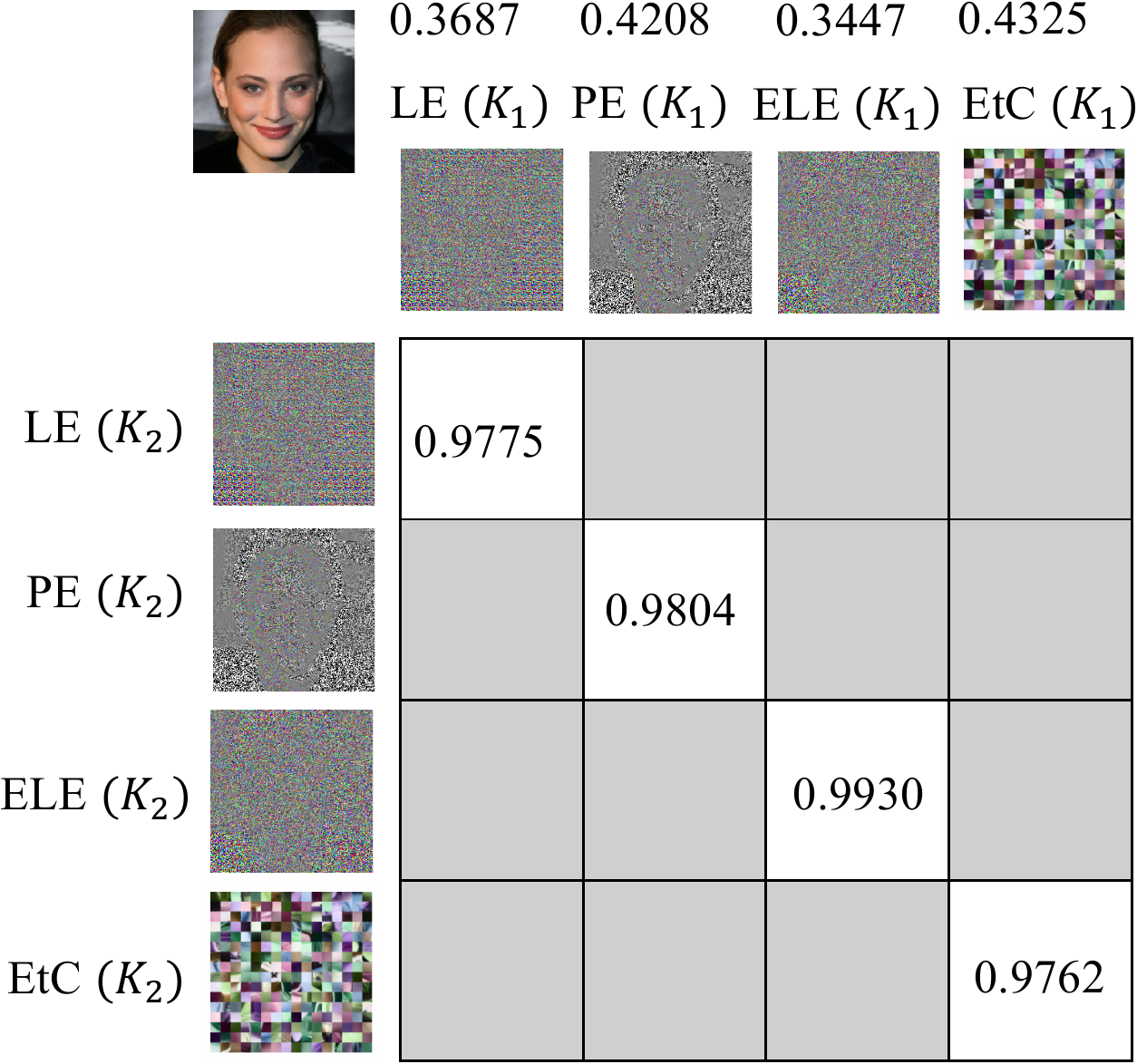}
\caption{CLIP similarity scores between plain-encrypted images and encrypted-encrypted images with different keys for four encryption methods. Labels near encrypted images indicate encryption algorithms and given keys. For example, LE ($K_1$) is used to label encrypted image by LE~\cite{2018-ICCETW-Tanaka} with key $K_1$, LE ($K_2$) is for key $K_2$, and so on.\label{fig:clip-score}}
\end{figure}

\textbf{Property 1 (Plain-Encrypted Similarity):} There is some similarity between encrypted images and plain ones, i.e.,
\begin{equation}
 \text{CLIP}(x, x') > 0.
\end{equation}
We calculated CLIP similarity scores~\cite{radford2021learning}, and the results are presented in Fig.~\ref{fig:clip-score}. PE~\cite{2019-Access-Warit} and EtC~\cite{kurihara2015encryption} have higher CLIP scores. PE is a special case of EtC, where the block size is 1, and only negative/positive transformation (channel-wise) and color component shuffling are applied. EtC is designed to preserve more color information for a compressibility trade-off. We utilize this property as conditional information for the generative model in the proposed attack.

\textbf{Property 2 (Encrypted-Encrypted Similarity):} An image encrypted with key $K_1$ ($x'_{K_1}$) is very close to that with key $K_2$ ($x'_{K_2}$). In other words, the CLIP similarity score of two encrypted images (by $K_1$ and $K_2$) is often more than $0.9$, i.e.,
\begin{equation}
 \text{CLIP}(x'_{K_1}, x'_{K_2}) > 0.9.
\end{equation}
This property is true for all learnable encryption methods. A CLIP similarity score matrix is given in Fig.~\ref{fig:clip-score}. Property 2 allows us to use any key for creating a synthetic image-encrypted paired dataset.

\begin{figure}[!t]
\centering
\includegraphics[width=\linewidth]{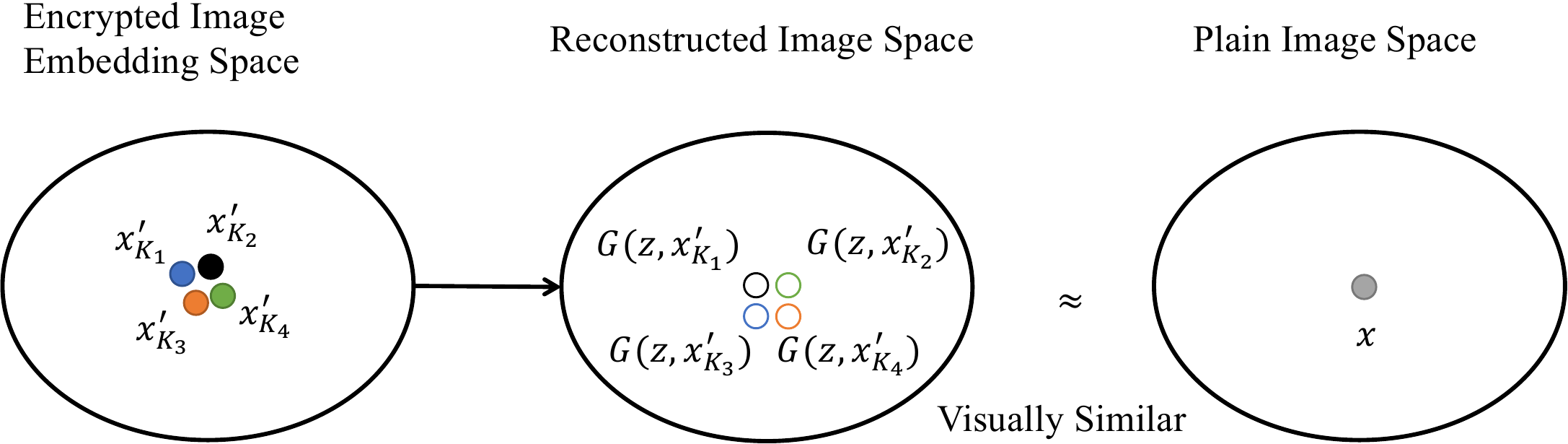}
\caption{Proposed attack assumption. We assume that images encrypted with different keys are close to one another in CLIP embedding space; thus, encrypted image embedding can effectively be used as conditional information to generative models.\label{fig:assumption}}
\end{figure}

Therefore, we assume that different images encrypted with different keys (i.e., ${x'_{K_1}, x'_{K_2}, x'_{K_3}, x'_{K_4}, \ldots}$) are close to one another in the CLIP embedding space as shown in Fig.~\ref{fig:assumption}. Given an encrypted image $x'$ as conditional information, a generative model $G$ is able to generate a visually similar image that corresponds to a plain image $x$ from pure Gaussian noise $z$ as follows:
\begin{equation}
 G(z, x') \sim x.
\end{equation}

\subsection{Overview}
\begin{figure}[!t]
\centering
\includegraphics[width=\linewidth]{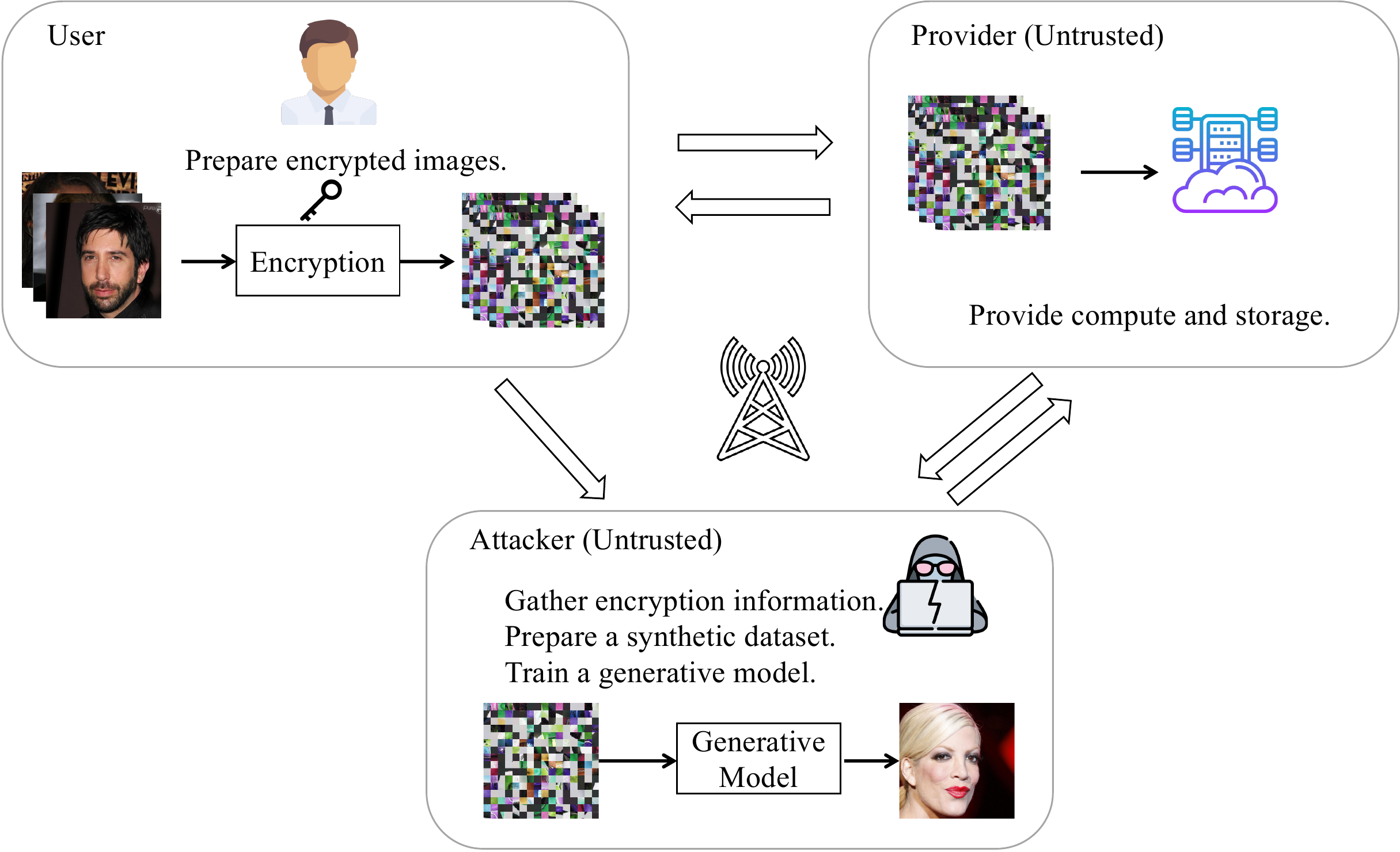}
\caption{Scenario of learnable image encryption exposed to attacks.\label{fig:overview-flow}}
\end{figure}

We consider the scenario in Fig.~\ref{fig:overview-flow}, where a user, a cloud provider, and an attacker may be involved. The user prepares encrypted images and trains/tests a classification model by using encrypted images at an untrusted cloud provider. The attacker may know the underlying encryption algorithm and carries out an attack to recover visual information from encrypted images.

The proposed method can be implemented in two ways: a StyleGAN-based method and diffusion-based method, denoted as Style attack and LDM attack, respectively.

\begin{figure}[t]
\centering
\subfloat[]{\includegraphics[width=\linewidth]{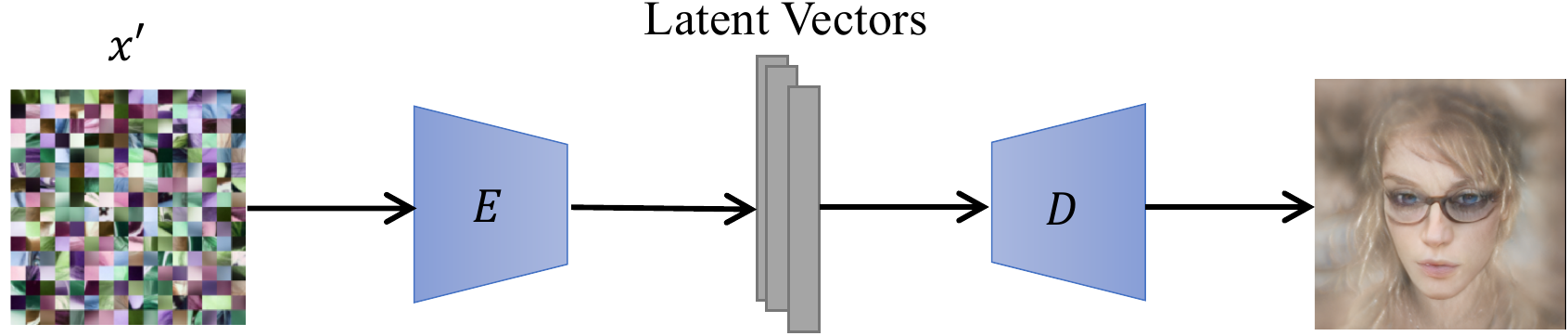}%
\label{fig:attack1}}
\hfil
\subfloat[]{\includegraphics[width=\linewidth]{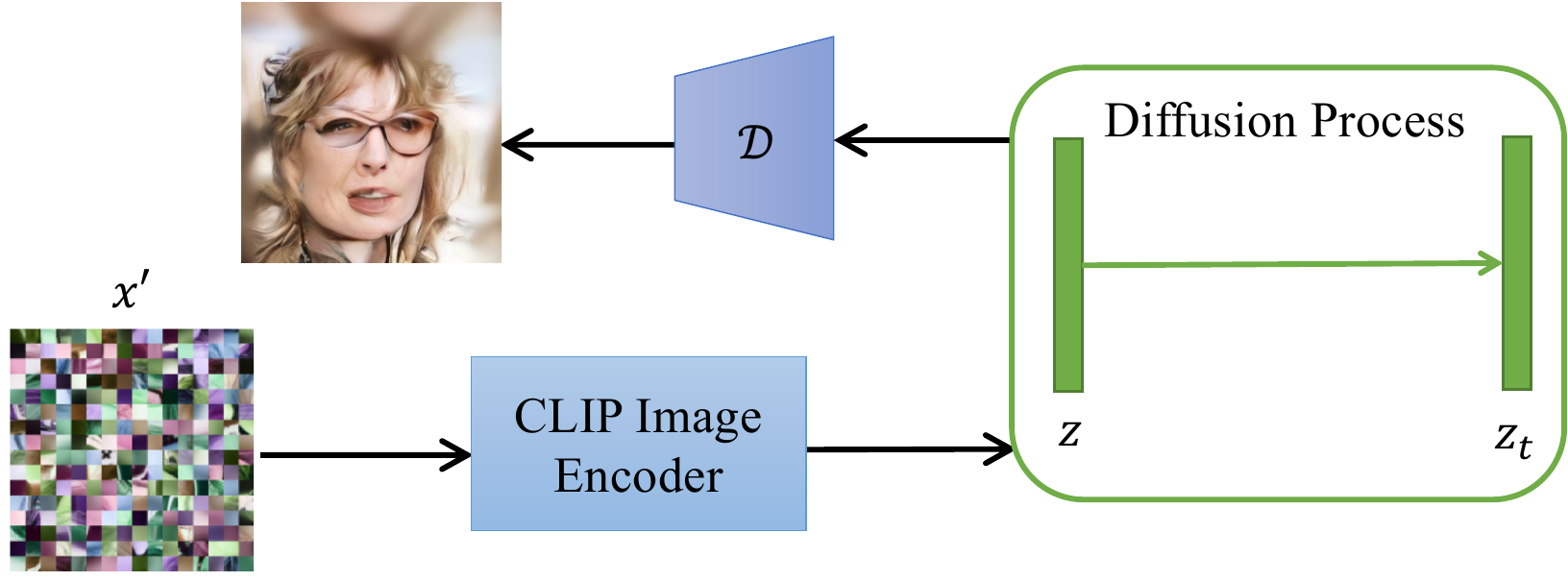}%
\label{fig:attack2}}
\caption{Overview of proposed generative model-based attack that recovers visual features from encrypted images. (a) Style attack. (b) LDM attack.\label{fig:overview}}
\end{figure}

\subsection{Style Attack}
Figure~\ref{fig:attack1} shows an overview of the Style attack. It utilizes an encoder $E$ that is a StyleGAN encoder~\cite{richardson2021encoding} and the pre-trained StyleGAN2 decoder $D$ so that $E$ extracts styles from an encrypted image $x'$, and the extracted styles are used to generate a plausible image with the decoder $D$~\cite{karras2020analyzing}, i.e.,
\begin{equation}
 D(z, E(x')) \sim x,
\end{equation}
where $z$ is constant noise input of the StyleGAN2 generator, and $x$ is the original plain image. This method relies heavily on the specific pre-trained StyleGAN2 generator.

\subsubsection{Training}
We train the StyleGAN encoder~\cite{richardson2021encoding} by using pairs of plain images and encrypted ones. As shown in Fig.~\ref{fig:attack1}, encoder $E$ encodes encrypted images to latent vectors that are combined with an average latent vector from decoder $D$. Then, decoder $D$ generates plausible images from the latent vectors. During training, we schedule a random key every epoch so that the encoder can generalize to extract styles from an encrypted image with any key. Note that random key scheduling plays an important role in the Style attack. In addition, as in the StyleGAN encoder~\cite{richardson2021encoding}, decoder $D$ is frozen during the training.

\subsubsection{Inference}
To execute the Style attack, trained encoder $E$ extracts styles from an encrypted image with any key and provides latent vectors. Decoder $D$, which is the pre-trained StyleGAN2 decoder, takes the resulting vectors and generates a plausible image that contains some visual information from a plain image.

\subsection{LDM Attack}
While following the same idea of recovering some visual features from encrypted images as in the Style attack, we also implement the proposed method with a state-of-the-art latent diffusion model. Figure~\ref{fig:attack2} depicts an overview of the LDM attack. This method is a guided LDM with CLIP image embedding, and $\mathcal{D}$ is the decoder of a general-purpose autoencoder, unlike the StyleGAN2 decoder in the Style attack (Fig.~\ref{fig:attack1}). Generally, there are three ways to condition a diffusion model: input concatenation~\cite{saharia2022image}, denormalization (modulation)~\cite{dhariwal2021diffusion}, and cross-attention~\cite{nichol2021glide}. We use denormalization conditioning in the proposed attack. Inspired by text-to-image models, GLIDE~\cite{nichol2021glide} and stable diffusion~\cite{rombach2022high}, we utilize the OpenAI CLIP ViT-L/14~\cite{radford2021learning} image encoder to condition the model on encrypted images. The CLIP image encoder produces a 768-dimensional embedding vector for a given encrypted image. This embedding is projected to a linear layer to form shift and scale factors, and the resulting factors are used to denormalize every ResNet block in the U-Net backbone of the diffusion model.

\subsubsection{Training}
\begin{figure}[!t]
\centering
\includegraphics[width=\linewidth]{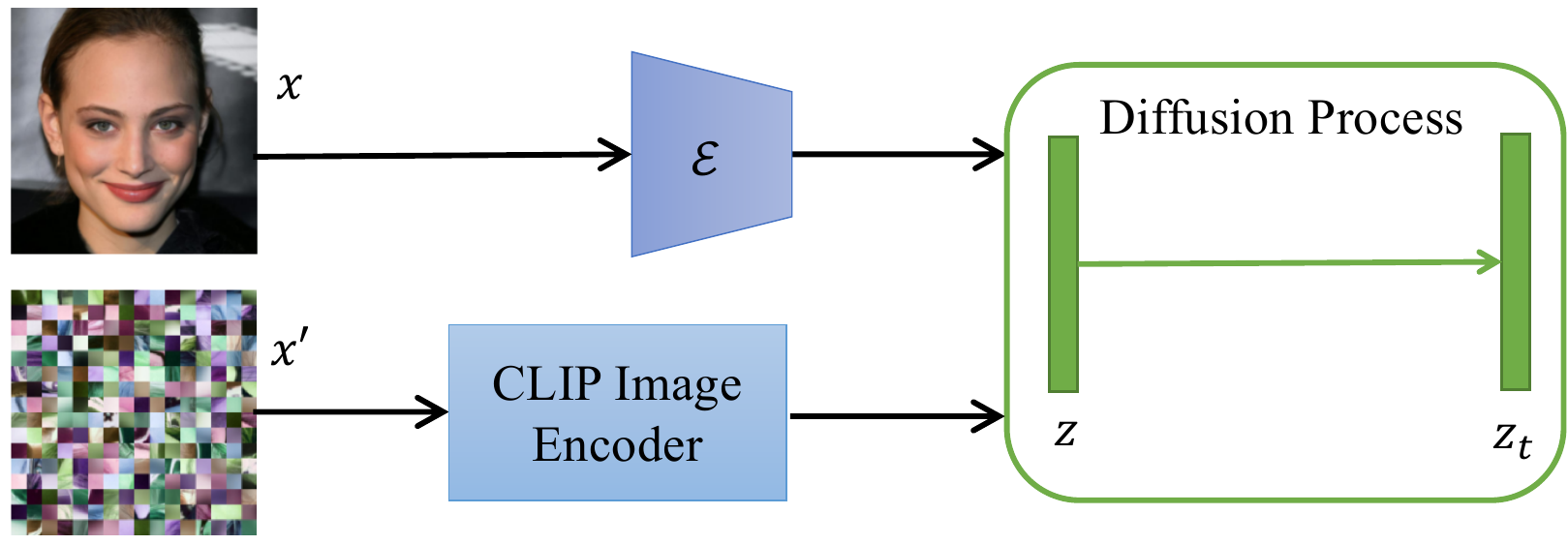}
\caption{High-level overview of proposed LDM attack training.\label{fig:forward-pass}}
\end{figure}

First, we prepare a training plain-encrypted paired dataset, $\{(x, x')\}_{x \in X}$, where $X$ is a plain image dataset, $x$ is a plain image, and $x'$ is a corresponding encrypted image. The proposed method adopts LDM~\cite{rombach2022high}. Therefore, the pre-trained autoencoder's encoder $\mathcal{E}$ is required during training. As illustrated in Fig.~\ref{fig:forward-pass}, plain-image $x$ is encoded by $E$ to a 2D latent representation, and cipher-image $x'$ is encoded by the CLIP image encoder for conditioning the ResNet block of the U-Net backbone in the LDM model. We omit the detailed architecture of LDM for simplicity. Interested readers are encouraged to refer to the original LDM paper~\cite{rombach2022high}. In addition, to better support classifier-free guidance sampling~\cite{ho2022classifier}, we randomly discard \SI{10}{\percent} of conditioning during training.

\subsubsection{Sampling}
To perform the proposed attack, we use classifier-free guidance sampling~\cite{ho2022classifier} that enables better quality sampling for the trained LDM.\@ During sampling, the output of the model with classifier-free guidance is
\begin{equation}
 \hat{\epsilon}_\theta(x_t|x') = \epsilon_\theta(x_t|\emptyset) + s \cdot (\epsilon_\theta(x_t|x') - \epsilon_\theta(x_t|\emptyset)),
\end{equation}
where $s > 1$ is a guidance scale, and $\emptyset$ is a null condition. For implementation, we use null condition $\emptyset$ as zeros. As illustrated in Fig.~\ref{fig:attack2}, the proposed method can recover visual information by sampling the trained LDM with CLIP image embedding and classifier-free guidance.

\subsection{Evaluation Metric\label{sec:lpips}}
We assess the perceptual quality of images reconstructed by the proposed attack by using the learned perceptual image patch similarity (LPIPS) metric~\cite{zhang2018unreasonable}. This metric is well known and widely used as a perceptual loss in image generation tasks. The previous attack also utilizes the LPIPS metric for evaluating the perceptual information of reconstructed images~\cite{madono2021gan}. A higher LPIPS score means two images are further, and lower means two images are similar.

The LPIPS score between a plain-image $x$ and a cipher-image $x'$ is calculated by using extracted features from $L$ layers of a pre-trained network as~\cite{zhang2018unreasonable}:
\begin{equation}
d(x',x) = \sum_l \dfrac{1}{H_l W_l} \sum_{h,w} || w_l \odot ( \hat{y}_{ehw}^l - \hat{y}_{hw}^l ) ||_2^2,
\end{equation}
where $H_l$ and $W_l$ are a spatial dimension of a feature map, $w_l$ is a scaling vector, $\odot$ is an element-wise multiplication operation, and $\hat{y}_{hw}^l$ and $\hat{y}_{ehw}^l$ are corresponding extracted feature maps of the $l$\textsuperscript{th} layer.

\section{Experiments and Discussion\label{sec:experiment}}
\subsection{Setup}
We used the CelebA-HQ dataset~\cite{karras2018progressive}, which is a high-quality version of the CelebA dataset~\cite{liu2015faceattributes}. The dataset consists of 30,000 male and female face images, where 28,000 images were used for training, and 2,000 images were reserved for testing. To further evaluate the proposed attack for scalability, we also utilized the ImageNet~\cite{ILSVRC15} dataset. ImageNet comprises 1.28 million color images for training and 50,000 color images for validation. We resized images to a dimension of $256 \times 256$.

For LDM implementation, we modified a simplified version\footnote{https://github.com/JD-P/cloob-latent-diffusion} of the original LDM implementation. We trained the LDM for $80,000$ steps for all experiments with a fixed learning rate value of $0.00003$. We used the AdamW optimizer~\cite{loshchilov2018decoupled} with a decay value of $0.01$. For attacking EtC images, we trained the LDM in two stages. In the first stage, the LDM was trained with images encrypted by only the block scrambling encryption step. Then, the LDM was trained with full EtC images in the second stage. Each stage took $80,000$ steps.

\subsection{Results for Celeba-HQ}
We trained LDM models (LDM attack) and StyleGAN encoder models (Style attack) for attacking four recent learnable encryption methods: EtC~\cite{kurihara2015encryption}, PE~\cite{2019-Access-Warit}, LE~\cite{2018-ICCETW-Tanaka}, and ELE~\cite{madono2020block}. We subjectively visualize the reconstructed images.

\subsubsection{EtC}
Figure~\ref{fig:res-etc} shows images reconstructed by the proposed attack. The reconstructed images for EtC showed identifiable information such as skin color, gender, beard, eye glasses, etc. We confirmed that images encrypted with different keys also revealed similar visual information.

\subsubsection{PE}
As shown in Fig.~\ref{fig:res-pe}, the proposed Style attack reconstructs photorealistic images that contain similar styles as in plain images. However, the proposed LDM attack could not reveal much visual information, so the subjective quality of the reconstructed images were lower than that of the style attack.

\subsubsection{LE}
\begin{figure}[t]
\centering
\subfloat[]{\includegraphics[width=0.98\linewidth]{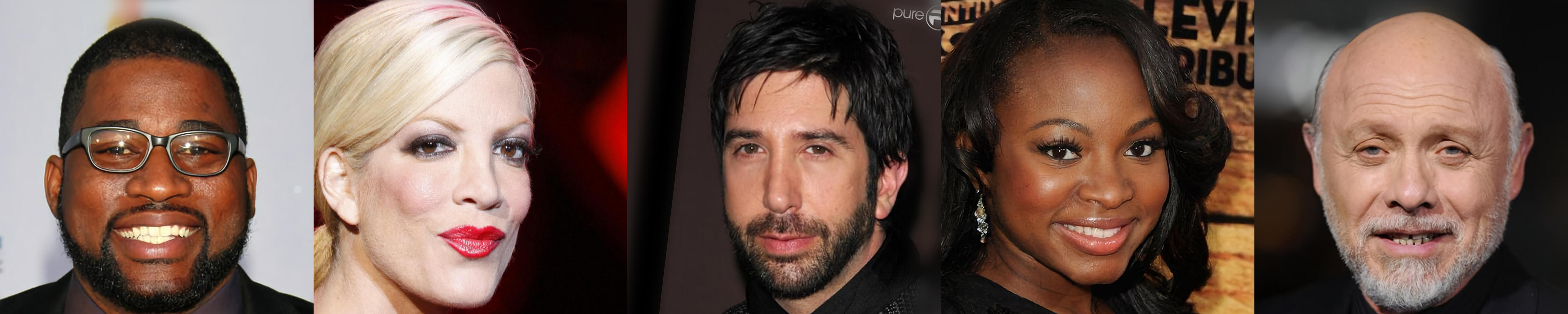}%
\label{fig:res-plain}}
\hfil
\subfloat[]{\includegraphics[width=0.49\linewidth]{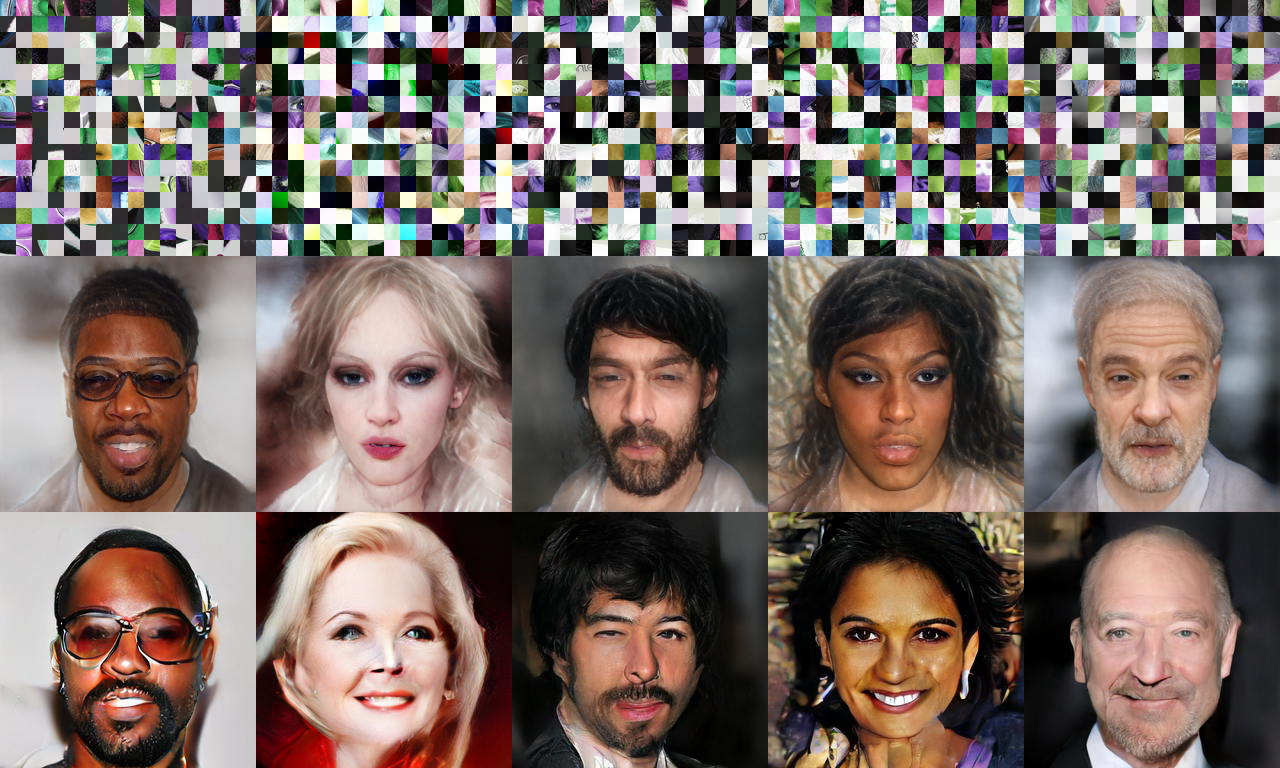}%
\label{fig:res-etc}}
\hfil
\subfloat[]{\includegraphics[width=0.49\linewidth]{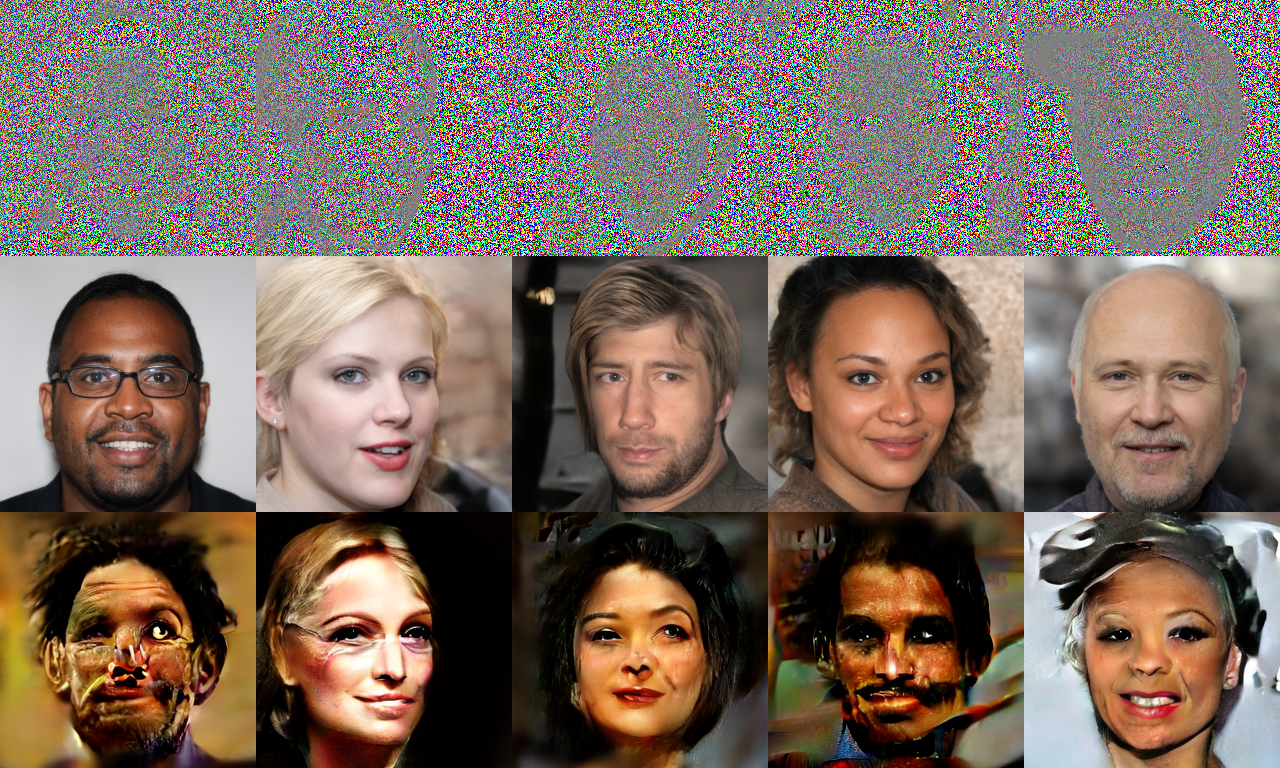}%
\label{fig:res-pe}}
\hfil
\subfloat[]{\includegraphics[width=0.49\linewidth]{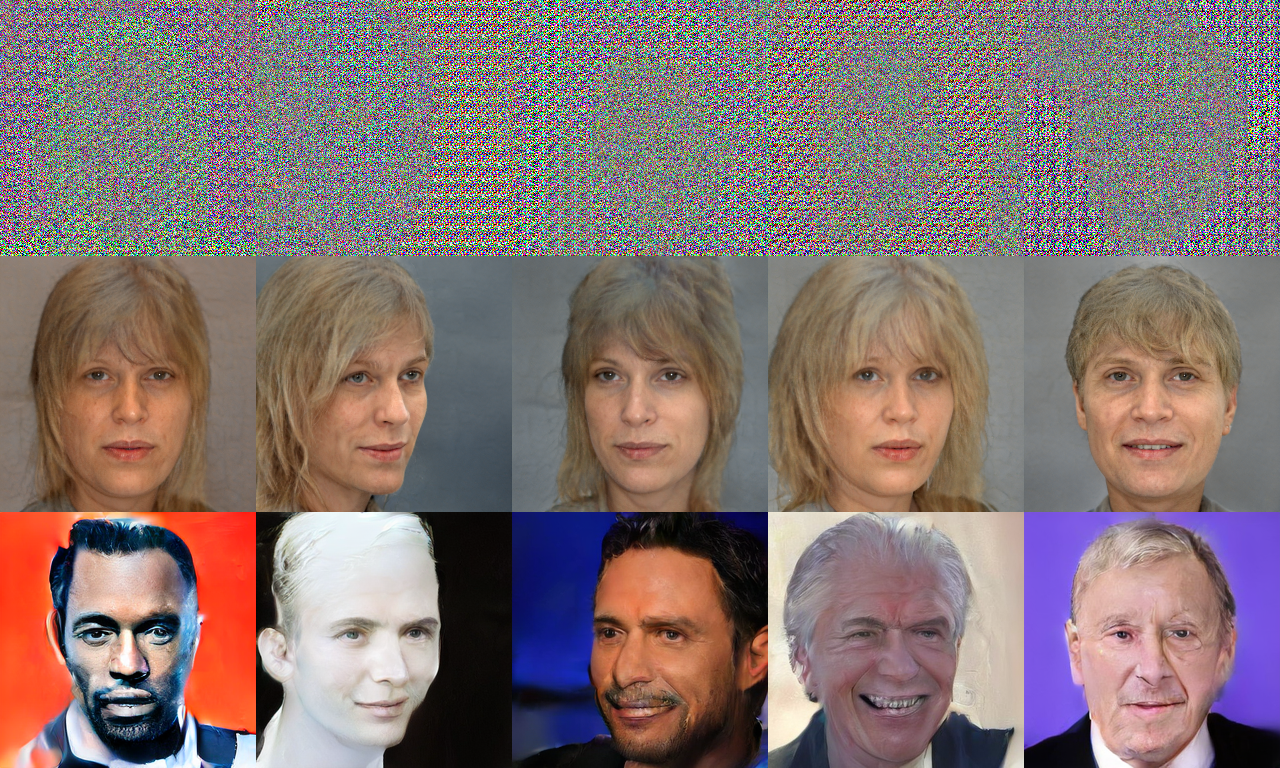}%
\label{fig:res-le}}
\hfil
\subfloat[]{\includegraphics[width=0.49\linewidth]{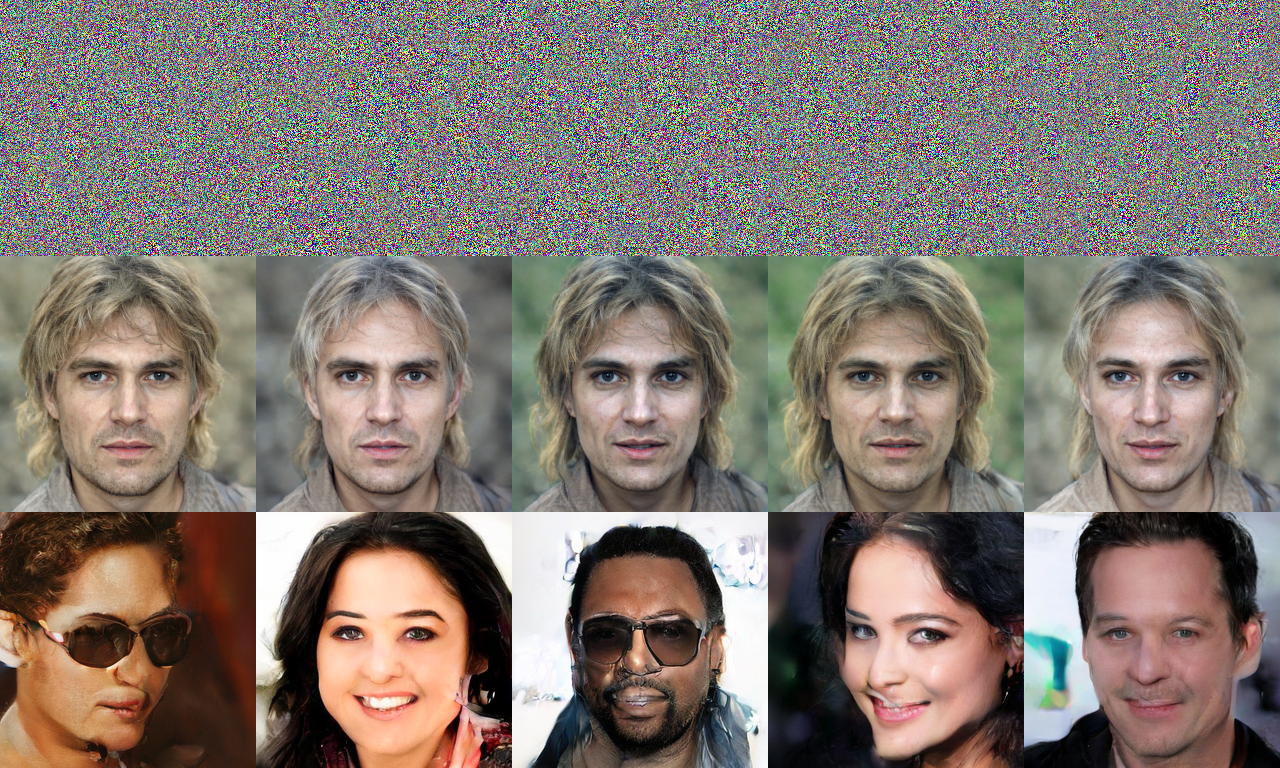}%
\label{fig:res-ele}}
\hfil
\caption{Reconstructed images by proposed generative attack. First row displays encrypted images, second row shows images reconstructed by Style attack, and third row displays images reconstructed by LDM attack. (a) Plain images. (b) EtC~\cite{kurihara2015encryption} images and their reconstruction. (c) PE~\cite{2019-Access-Warit} images and their reconstruction. (d) LE~\cite{2018-ICCETW-Tanaka} images and their reconstruction. (e) ELE~\cite{madono2020block} images and their reconstruction.\label{fig:res}}
\end{figure}

Although LE does not apply a block scrambling encryption step, the proposed attack cannot reconstruct visual information accurately. As shown in Fig.~\ref{fig:res-le}, the images reconstructed by the LDM attack revealed some visual features, but some of the features were false positives. In contrast, the Style attack did not have any identifiable facial features. Therefore, the proposed attack is not suitable for LE~\cite{2018-ICCETW-Tanaka}.

\subsubsection{ELE}
For the ELE method, the proposed generative attack (both LDM and Style attack) did not work well. The reason is when using a block size of $16 \times 16$, ELE destroys most of the local information. Therefore, the Style attack can generate only an average face, and the conditioning information for guiding the LDM is not enough.

\subsection{Facial Attribute Classification}
\begin{figure}[!t]
\centering
\includegraphics[width=\linewidth]{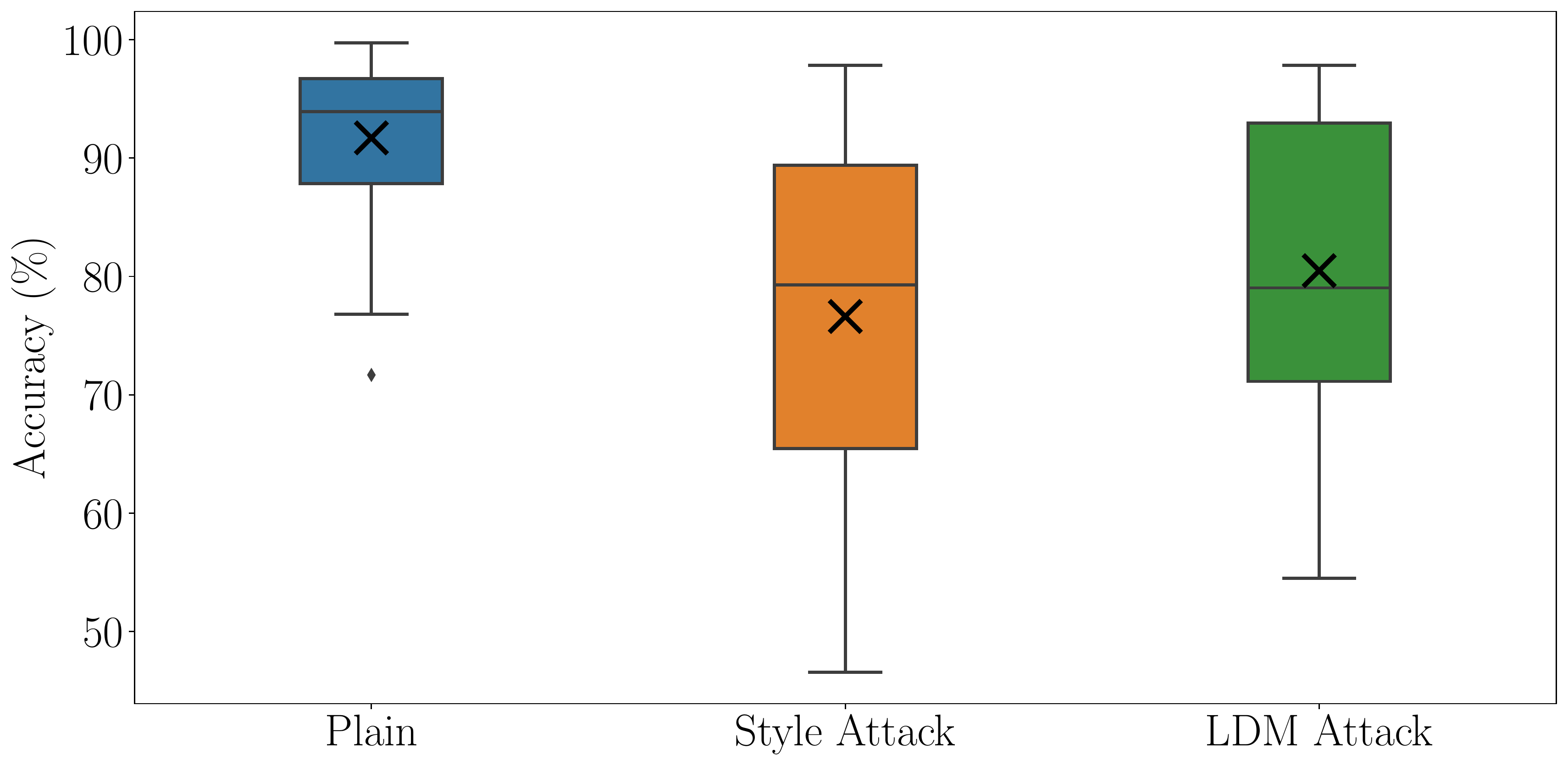}
\caption{Overall accuracy (\SI{}{\percent}) of 40 face attribute classification for plain and reconstructed images. Accuracy was calculated over test set (19,962) images. Boxes span from first to third quartile, referred to as $Q_1$ and $Q_3$, and whiskers show maximum and minimum values in range of $[Q_1 - 1.5(Q_3 - Q_1), Q_3 + 1.5(Q_3 - Q_1)]$. Band and cross inside boxes indicate median and average values, respectively. Dots represent outliers.\label{fig:faceattr}}
\end{figure}

\begin{figure*}[!t]
\centering
\includegraphics[width=\linewidth]{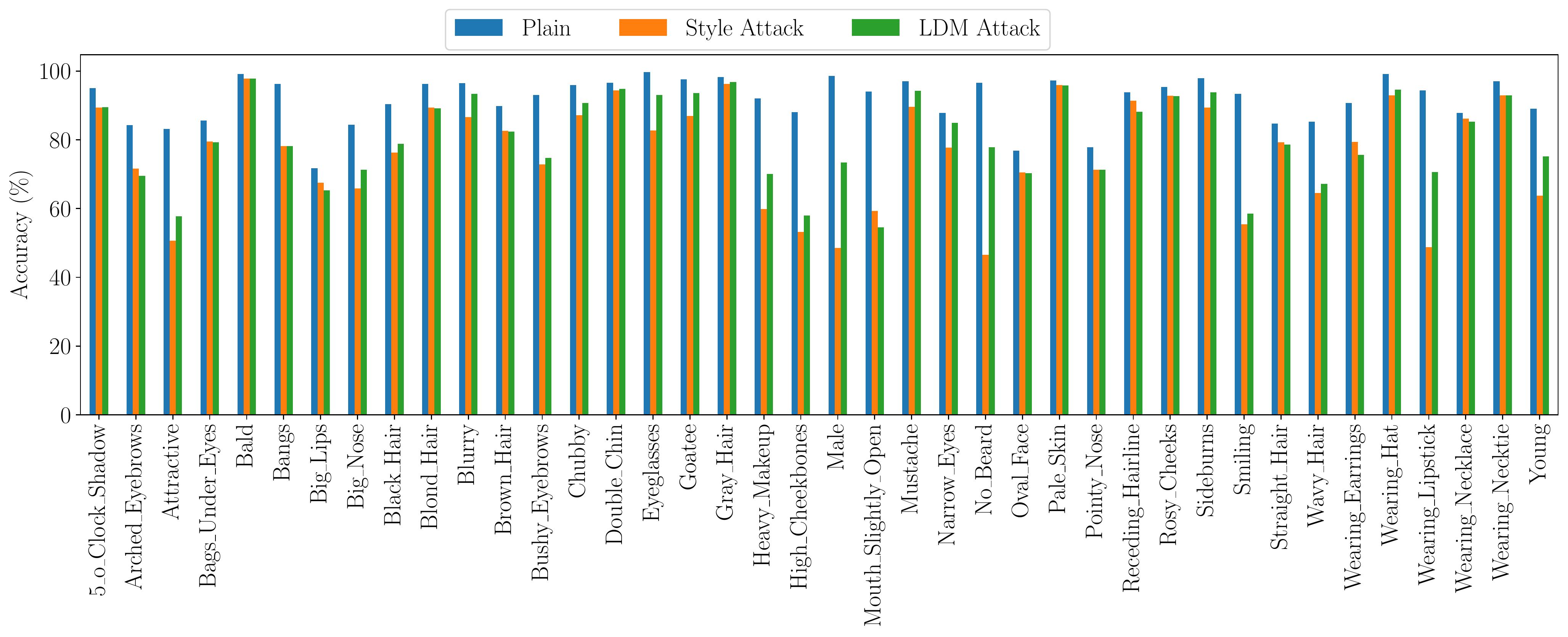}
\caption{Facial attribute classification for plain and reconstructed images by both Style and LDM attacks.\label{fig:classification}}
\end{figure*}

In addition, we carried out an additional experiment with the CelebA~\cite{liu2015faceattributes} dataset, which has $40$ face attribute annotations. We utilized a finetuned MobileNetV2~\cite{sandler2018mobilenetv2} (pre-trained on ImageNet) with 40 multi-head binary classifiers to classify facial attributes. To evaluate how much we could identify a person from a reconstructed image, we classified both plain images and reconstructed images. The results are plotted in Fig.~\ref{fig:faceattr}, where the whole test set ($19962$ face images) of CelebA was used to calculate the classification accuracy. From the results, the overall accuracy of plain images was \SI{91.71}{\percent}, that of the Style attack was \SI{76.62}{\percent}, and that of the LDM attack was \SI{80.49}{\percent}. Although the Style attack provided a higher LPIPS score, the LDM attack had superior performance on facial attribute classification. In addition, Fig.~\ref{fig:classification} further shows the classification performance for each face attribute. Therefore, the results demonstrated that images reconstructed by the proposed attack reveal some visually identifiable information.

\subsection{Results for ImageNet}
\begin{figure}[!t]
\centering
\includegraphics[width=\linewidth]{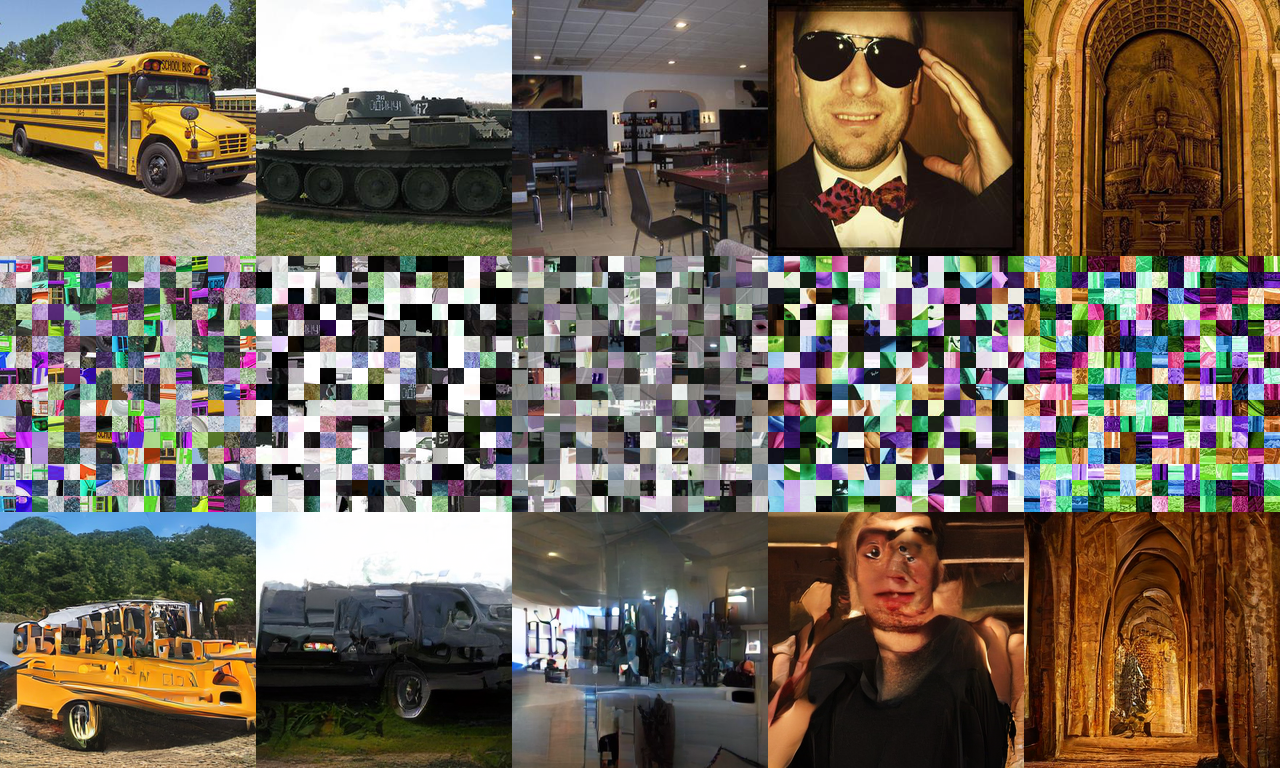}
\caption{Images reconstructed images by proposed LDM attack for EtC~\cite{kurihara2015encryption} with ImageNet dataset. First row is for plain images, second row is for encrypted images, and third row is for reconstructed images.\label{fig:res-imagenet}}
\end{figure}

To further analyze the performance of the proposed generative attack, we trained an LDM attack on the ImageNet dataset for the EtC scheme. We were not able to evaluate the Style attack because there is no available pre-trained StyleGAN2 model for ImageNet. Figure~\ref{fig:res-imagenet} shows random images from the ImageNet dataset with corresponding encrypted images and reconstructed ones. From the figure, the LDM attack in its current form revealed visual styles even for natural images. However, the attack could not reconstruct detailed visual information.

\subsection{Comparison with State-of-the-Art Attacks}
\begin{figure}
 \centering
 \begin{tabular}{cc}
 \parbox[t]{1mm}{\rotatebox[origin=c]{90}{Plain}} &
 \begin{minipage}{\linewidth}
 \centering
 \includegraphics[width=\linewidth]{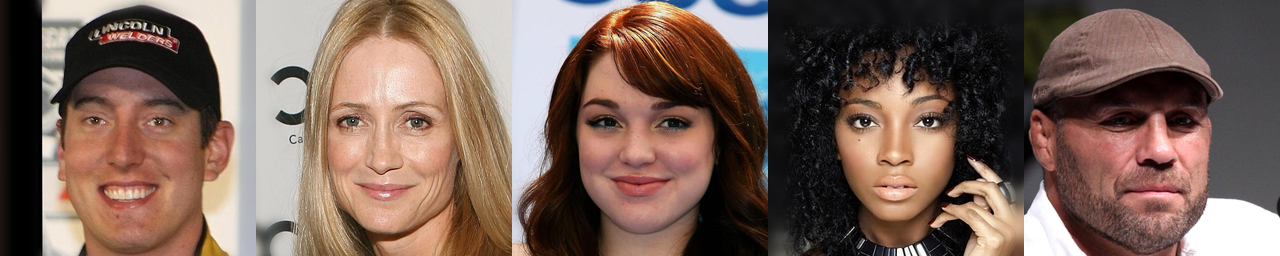}
 \end{minipage} \\
 \parbox[t]{1mm}{\rotatebox[origin=c]{90}{EtC}} &
 \begin{minipage}{\linewidth}
 \centering
 \includegraphics[width=\linewidth]{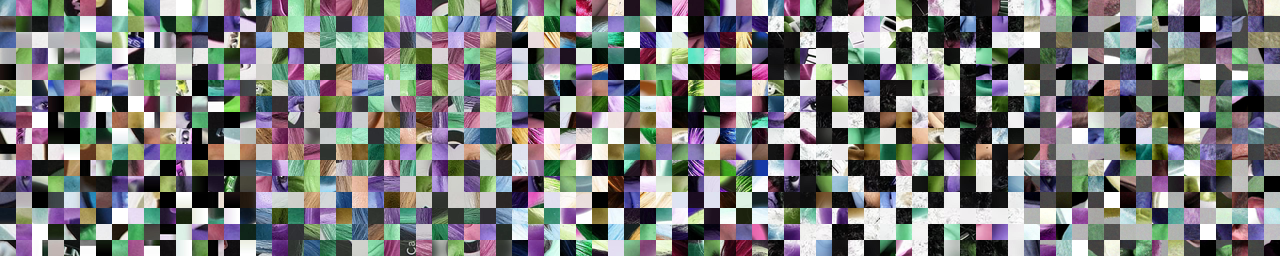}
 \end{minipage} \\
 \parbox[t]{1mm}{\rotatebox[origin=c]{90}{FR~\cite{chang2020attacks}}} &
 \begin{minipage}{\linewidth}
 \centering
 \includegraphics[width=\linewidth]{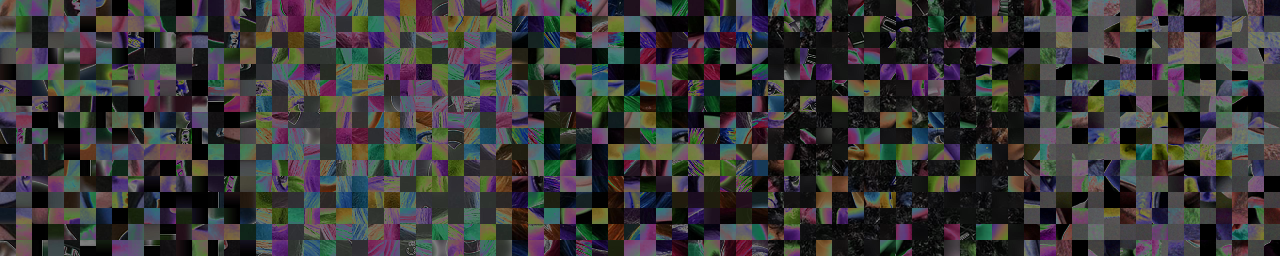}
 \end{minipage} \\
 \parbox[t]{1mm}{\rotatebox[origin=c]{90}{GAN~\cite{madono2021gan}}} &
 \begin{minipage}{\linewidth}
 \centering
 \includegraphics[width=\linewidth]{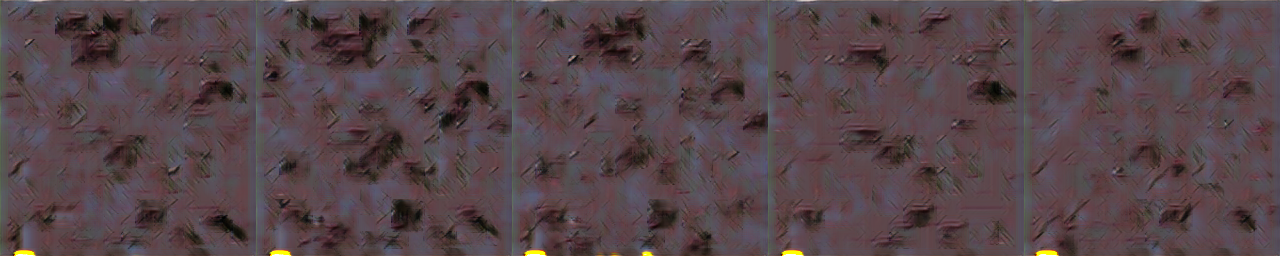}
 \end{minipage} \\
 \parbox[t]{1mm}{\rotatebox[origin=c]{90}{ITN~\cite{ito2021image}}} &
 \begin{minipage}{\linewidth}
 \centering
 \includegraphics[width=\linewidth]{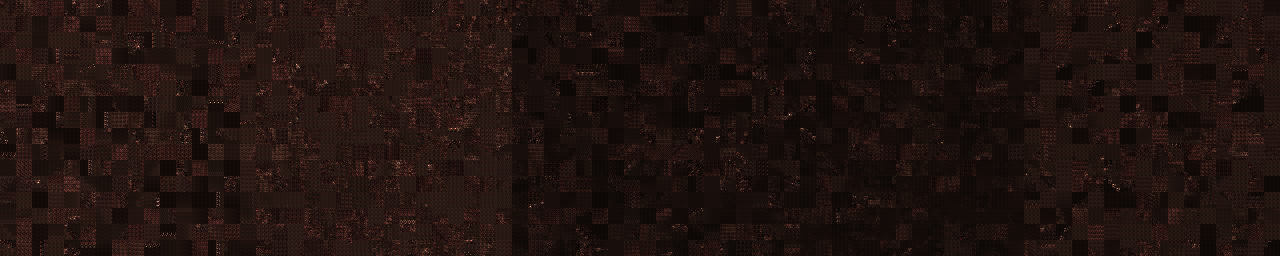}
 \end{minipage} \\
 \parbox[t]{1mm}{\rotatebox[origin=c]{90}{Style}} &
 \begin{minipage}{\linewidth}
 \centering
 \includegraphics[width=\linewidth]{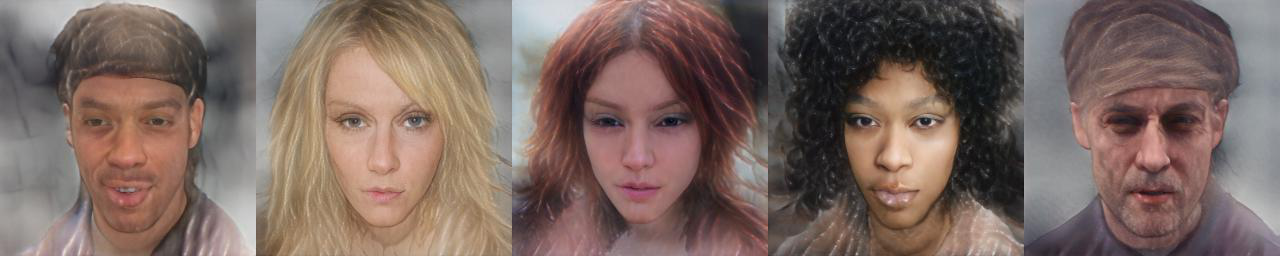}
 \end{minipage} \\
 \parbox[t]{1mm}{\rotatebox[origin=c]{90}{LDM}} &
 \begin{minipage}{\linewidth}
 \centering
 \includegraphics[width=\linewidth]{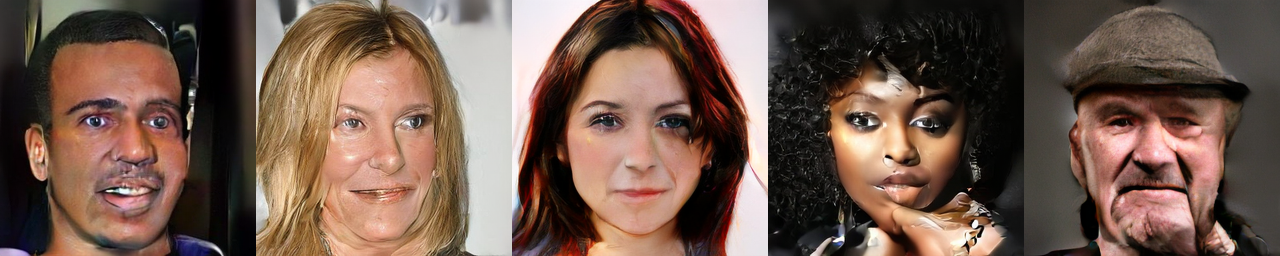}
 \end{minipage} \\
 \end{tabular}
\caption{Examples of plain, encrypted (EtC), and reconstructed images by different attacks.\label{fig:ex-comparison}}
\end{figure}

We focus on the EtC scheme because it contains a block scrambling encryption step and the previous attack in~\cite{madono2021gan} cannot recover visual information from EtC images. We compared the proposed attack with state-of-the-art attack methods: FR~\cite{chang2020attacks}, GAN attack~\cite{madono2021gan}, and ITN attack~\cite{ito2021image}. Examples of reconstructed images are shown in Fig.~\ref{fig:ex-comparison}, where conventional attacks could not recover any visual information from encrypted images. In contrast, both the proposed Style and LDM attacks revealed some visual features that were similar to the original plain images.

\begin{figure}[!t]
\centering
\includegraphics[width=\linewidth]{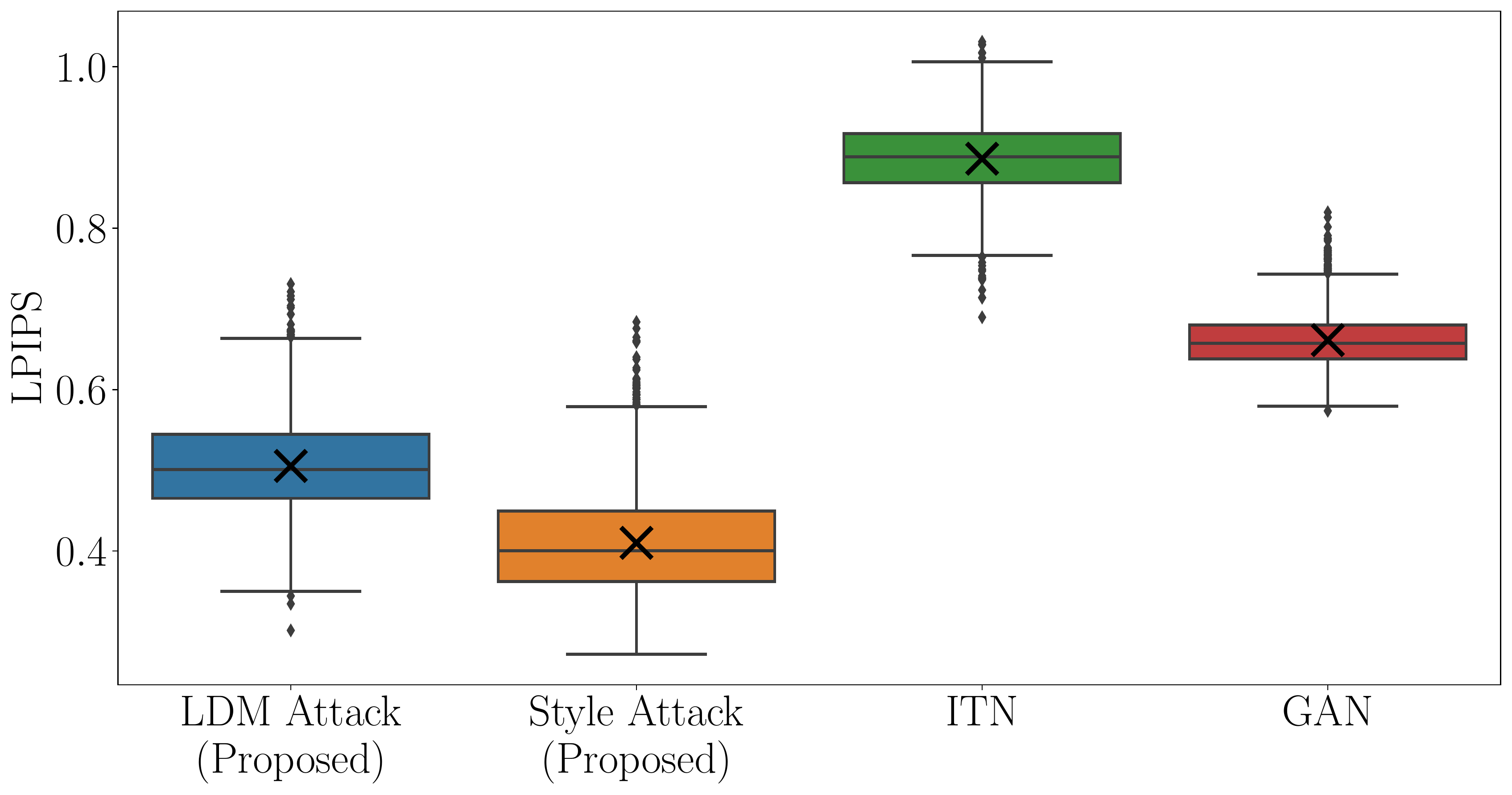}
\caption{LPIPS scores between plain and reconstructed images for proposed attack, ITN attack~\cite{ito2021image}, and GAN attack~\cite{madono2021gan}. Scores were calculated over validation set (2,000) images. Boxes span from first to third quartile, referred to as $Q_1$ and $Q_3$, and whiskers show maximum and minimum values in range of $[Q_1 - 1.5(Q_3 - Q_1), Q_3 + 1.5(Q_3 - Q_1)]$. Band and cross inside boxes indicate median and average values, respectively. Dots represent outliers.\label{fig:lpips}}
\end{figure}

Objectively, we also compared the proposed generative attack with previous attacks in terms of LPIPS scores. The LPIPS is the preferred way of measuring perceptual information between two image patches as described in Section~\ref{sec:lpips}. We used 2,000 images from the validation set (which was not included in the training), and the results are plotted in Fig.~\ref{fig:lpips}. The proposed method achieved smaller LPIPS scores, meaning that the resulting reconstructed images and plain images were perceptually similar. In contrast, the images reconstructed from the ITN attack~\cite{ito2021image} and GAN attack~\cite{madono2021gan} were perceptually different, indicated by the higher LPIPS scores. Therefore, from the LPIPS results, the proposed method outperformed the previous attack methods for reconstructing EtC images.

\subsection{Discussion and Limitations}
\subsubsection{Difference between Style and LDM attack}
There are two implementations of the proposed generative model-based attack: the Style and LDM attack. They are fundamentally different.
\begin{itemize}
\item The Style attack is a GAN-based method, and the LDM attack is a diffusion-based method.
\item In the Style attack, the encoder is trained, and the encrypted image is directly used as an input to the encoder during training. Therefore, it is more like an image-to-image translation approach even though the generator generates an image from noise and latent vectors. In contrast, the encoder is not trained in LDM.\@ Instead, LDM utilizes a frozen CLIP image encoder to generate an image embedding that is used as conditional information via demodulation.
\item Since the encoder is trained, the Style attack produces more photorealistic images. However, the Style attack relies on a pre-trained StyleGAN2 decoder, and the attack model cannot be trained without the available pre-trained decoder. In contrast, LDM can be trained on any dataset.
\end{itemize}

\subsubsection{Limitations}
\begin{figure}[!t]
\centering
\includegraphics[width=\linewidth]{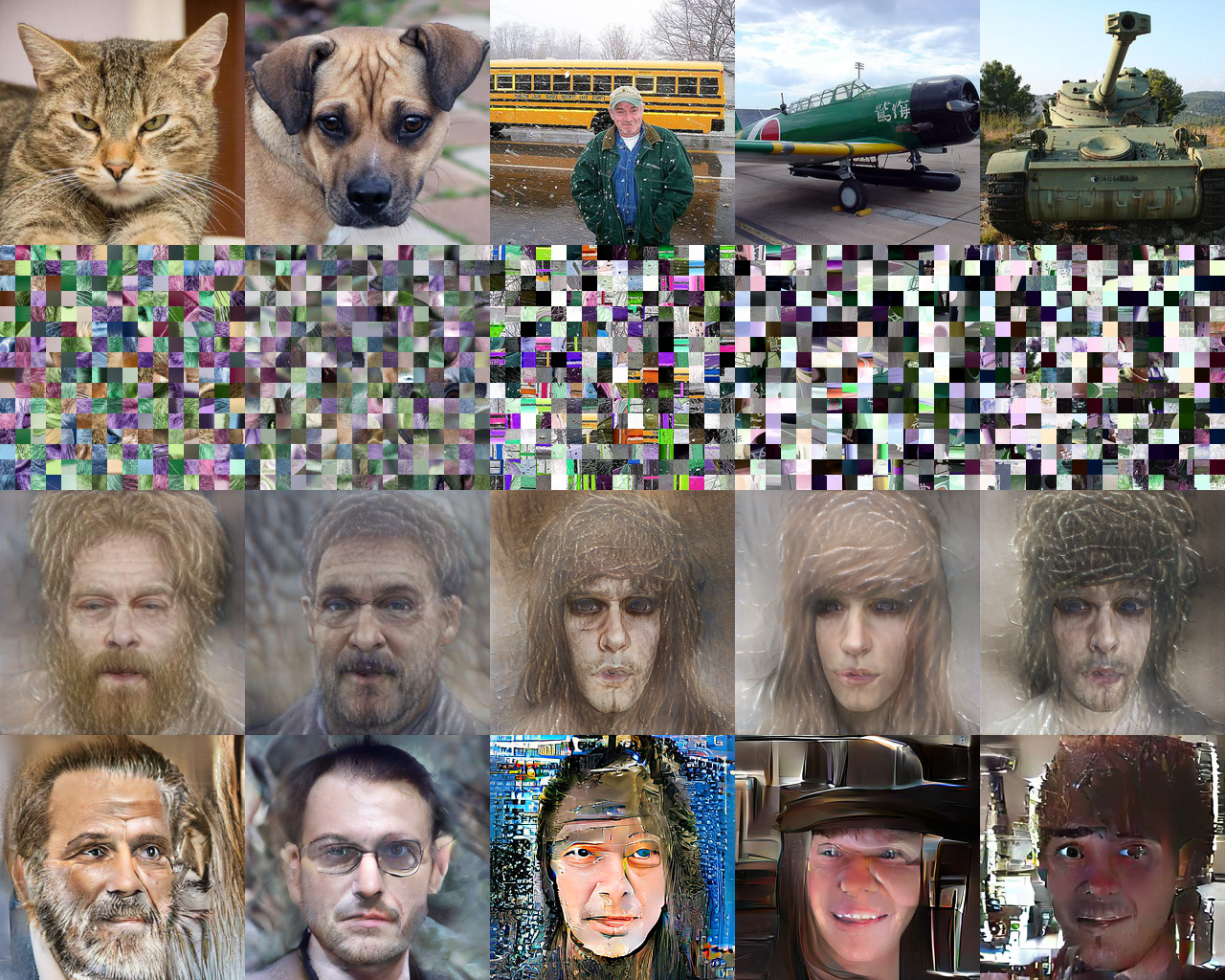}
\caption{Out-of-distribution reconstructed examples. First row is for plain images, second row is for encrypted images, third is for images reconstructed by Style Attack, and fourth is for images reconstructed by LDM attack.\label{fig:res-out}}
\end{figure}

The proposed attack shows that it is possible to extract styles from EtC images even though a block scrambling step is included in the encryption. However, there are certain limitations.
\begin{itemize}
\item The proposed attack cannot reconstruct an encrypted image as an identical plain image. It can only recover some styles from an encrypted image and generate a plausible image.
\item The proposed Style attack is dataset-specific and relies on a pre-trained StyleGAN2 generator. In contrast, the LDM attack is not specific to a dataset. However, the generative models trained on face datasets cannot be used to reconstruct images other rather than face images as shown in Fig.~\ref{fig:res-out}. From the experiment results on ImageNet, the LDM attack still cannot recover much perceptual information if an image contains multiple objects or a complex scene. Therefore, the proposed attack in its current form is limited to only recovering some styles from encrypted images. We shall conduct future research to improve the proposed generative model-based attack.
\end{itemize}

\section{Conclusion\label{sec:conclusion}}
In this paper, we proposed a generative model-based attack for the first time for learnable image encryption that is designed for privacy-preserving deep learning. We demonstrated that encrypted images can be directly used as input to a StyleGAN encoder or embeddings of encrypted images as conditional information to latent diffusion models to recover visual information. Experiment results show that images reconstructed by the proposed attack have some perceptual similarities to plain images. In addition, face attribute classification further confirms that reconstructed images can also be classified for the majority of face attributes. Overall, the proposed attack outperforms the previous attacks, especially for EtC images. However, there are still some limitations, and we shall improve the performance of the proposed attack as our future work.

\vspace{-0.5cm}

\bibliographystyle{IEEEtran}
\bibliography{IEEEabrv,/Users/maung/Dropbox/refs}

\end{document}